\pdfoutput=1

\documentclass[11pt]{article}

\usepackage{emnlp2021}

\usepackage{times}
\usepackage{latexsym}

\usepackage[T1]{fontenc}

\usepackage[utf8]{inputenc}

\usepackage{microtype}

\usepackage{bm}
\usepackage[linesnumbered,ruled]{algorithm2e}
\usepackage{amsmath}
\usepackage{mathtools}
\usepackage{amssymb}
\usepackage{amsthm}
\usepackage{array}
\usepackage{color}
\usepackage{multirow}
\usepackage{comment}
\usepackage{wrapfig}
\usepackage{lipsum} 
\usepackage{soul}
\usepackage{graphicx}
\usepackage{subcaption}
\usepackage{booktabs}
\usepackage{enumitem}

\DeclareMathOperator*{\argmax}{argmax}

\title{Towards Variable-Length Textual Adversarial Attacks}

\author{Junliang Guo$^{1}$, Zhirui Zhang$^{2}$, Linlin Zhang$^{3}$, Linli Xu$^{1}$, \\
{\bf Boxing Chen$^{2}$, Enhong Chen$^{1}$, Weihua Luo$^{2}$}
\\
$^{1}$Anhui Province Key Laboratory of Big Data Analysis and Application, \\
School of Computer Science and Technology, \\
University of Science and Technology of China \\
$^{2}$Alibaba DAMO Academy \quad
$^{3}$Zhejiang University\\
{\tt guojunll@mail.ustc.edu.cn, \{linlixu,cheneh\}@ustc.edu.cn} \\
{\tt \{zhirui.zzr, zll240651, boxing.cbx, weihua.luowh\}@alibaba-inc.com}
}

\date{}

\begin{document}
\maketitle
\begin{abstract}
Adversarial attacks have shown the vulnerability of machine learning models,
however, it is non-trivial to conduct textual adversarial attacks 
on natural language processing tasks 
due to the discreteness of data. Most previous approaches conduct attacks with the atomic \textit{replacement} operation, which usually leads to fixed-length adversarial examples and therefore limits the exploration on the decision space.
In this paper, we propose variable-length textual adversarial attacks~(VL-Attack) and integrate three atomic operations, namely \textit{insertion}, \textit{deletion} and \textit{replacement}, into a unified framework, by introducing and manipulating a special \textit{blank} token while attacking.
In this way, our approach is able to more comprehensively find adversarial examples around the decision boundary and effectively conduct adversarial attacks.
Specifically, our method drops the accuracy of IMDB classification by $96\%$
with only editing $1.3\%$ tokens while attacking a pre-trained BERT model.
In addition, fine-tuning the victim model with generated adversarial samples can improve the robustness of the model without hurting the performance,
especially for length-sensitive models.
On the task of non-autoregressive machine translation,
our method can achieve $33.18$ BLEU score on IWSLT14 German-English translation, achieving
an improvement of $1.47$ over the baseline model.
\end{abstract}

\section{Introduction}
\label{sec:intro}
While achieving great successes in various domains, machine learning models have been found vulnerable to adversarial examples, i.e., the original inputs with small perturbations that are indistinguishable to human knowledge can fool the model and lead to incorrect results~\citep{goodfellow2014explaining,kurakin2016adversarial,zhao2017generating}. Adversarial attacks and defenses are able to improve the robustness and security~\citep{szegedy2013intriguing,madry2017towards} as well as to explore the interpretability of machine learning models~\citep{ribeiro2018semantically,tao2018attacks}.

\begin{figure}[tb]
\centering
\includegraphics[width=0.9\linewidth]{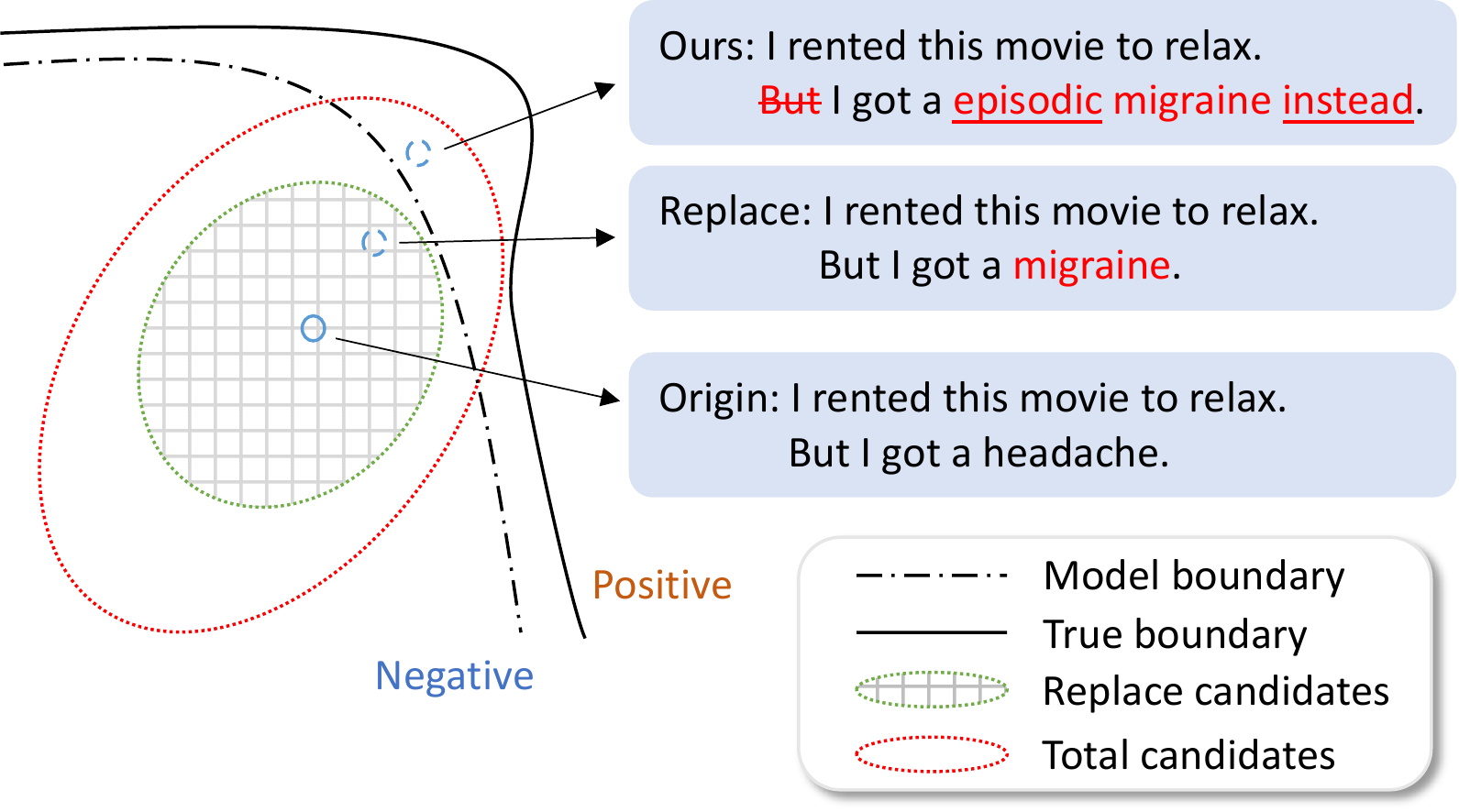}
\vspace{-10pt}
\caption{An illustration of the proposed VL-Attack method. 
Red, underline and strikeout tokens indicate replacements, insertions and deletions respectively.
Equipped with the proposed three atomic operations, our method is able to 
generate adversarial examples that are closer to the true decision boundary, which cannot be reached only by replacement.}
\label{fig:illus}
\vspace{-10pt}
\end{figure}

While having been extensively studied in computer vision models, adversarial attack on natural language processing models is more challenging as small perturbations of textual data may change its original semantic meaning and thus are perceptible. Most previous works utilize the \textit{replacement} operation to construct adversarial examples, including elaborately simulating natural noises such as typos by character-level replacement~\citep{ebrahimi2017hotflip,li2018textbugger} and word-level synonym replacement~\citep{papernot2016crafting,alzantot2018generating}, to ensure the generated examples will not alter the semantics. 
However, naturally, most adversarial examples could have different lengths from the original sentence, and the replacement operation is only able to generate fixed-length examples and thus
produce a limited subset of the total adversarial candidates. We provide an illustration in Figure~\ref{fig:illus}, where replacements cannot 
generate variable-length adversarial examples.

In fact, 
an original sentence can be converted to any adversarial candidate with the combination of three one-step atomic operations:
\textit{replacement}, \textit{insertion} and \textit{deletion}.
Inspired by this observation, we propose a paradigm of white-box variable-length adversarial attacks to comprehensively explore adversarial examples.
The whole attack procedure consists of multiple steps.
In each step, we conduct an attack utilizing an operation randomly sampled from the proposed three types of atomic operations.
Theoretically, our method is able to reach adversarial examples
with any word-level Levenshtein distance~\citep{levenshtein1966binary} to the inputs, therefore covering the full adversarial candidates set.

We propose tailored attacking methods for the proposed three atomic operations.
Specifically, we make \textit{insertion} and \textit{deletion} possible by converting them into gradient-based methods
with the introduction of a special blank token \texttt{[BLK]}.
Given a pre-trained victim model such as BERT, we introduce \texttt{[BLK]} while fine-tuning it on downstream tasks. By randomly inserting the blank token into inputs while keeping the target unchanged, the token is trained as a ``blank space'' to the model 
and its occurrence will not affect the original semantic information.
Then, we implement \textit{insertion} and \textit{deletion} with the help of the blank token. For insertion, we first insert a \texttt{[BLK]} token in a chosen position of the input. Then we choose the content to insert by calculating the 
similarities between the gradients of \texttt{[BLK]} and tokens in the vocabulary. While deleting, we evaluate the importance of all tokens by comparing with \texttt{[BLK]} in the gradient space, and then delete the most important token of the sentence.
And for replacement, we follow the gradient based method utilized in~\citep{ebrahimi2017hotflip,wallace2019universal} and replace 
a token with the one that maximizes the first-order approximation of the prediction loss.

Besides conducting attacks, our method can also be viewed as a data augmentation method to enhance the robustness and performance of models. 
Since the proposed method is able to generate variable-length adversarial examples, it is particularly beneficial to length-sensitive models, such as non-autoregressive neural machine translation (NAT)~\citep{gu2017non,guo2019non,guo2020fine} which requires 
predicting the target length before decoding.
Therefore, in experiments, we evaluate our method on two tasks including natural language understanding~(NLU) and NAT with mask-predict decoding~\citep{ghazvininejad2019mask}. 
We verify the effectiveness of the proposed method on NLU tasks, where our method successfully attacks the pre-trained BERT~\citep{devlin2018bert} model when fine-tuning on downstream tasks. Specifically, after the attack, the model's IMDB classification accuracy drops from $90.4\%$ to $3.5\%$ 
by only editing $1.3\%$ tokens.
Human evaluation verifies that the generated adversarial examples are highly accurate in grammatical, semantical and label preserving perspectives.
On NAT, we show that our method can be effectively used for data augmentation. By integrating the adversarial examples of training samples, 
our method boosts the performance of the mask-predict model on various machine translation tasks. Specifically, we achieve $33.18$ BLEU score on IWSLT14 German-English translation with
an improvement of $1.47$ over the baseline model.

Our main contributions can be summarized as follows:
\begin{itemize}[leftmargin=*]
    \setlength{\itemsep}{0pt}
    \setlength{\parskip}{0pt}
    \item We propose a variable-length adversarial attack method on textual data, which integrates three atomic operations, including replacement, insertion and deletion, into a unified framework.
    \item Our method successfully attacks the pre-trained BERT model on NLU tasks with higher attacking success rate and lower perturbation rate than previous approaches, while being semantic and label preserving from human judgement.
    \item Our method can be viewed as a data augmentation method which is specifically beneficial for length-sensitive models, and is able to achieve strong performance on NAT tasks.
\end{itemize}

\section{Related Work}
\label{sec:related}

\subsection{Textual Adversarial Attack}
\label{sec:related-text-attack}

Adversarial attack has been greatly studied in computer vision, and most works execute attack with gradient-based perturbation on the continuous space~\citep{szegedy2013intriguing,goodfellow2014explaining}.
For textual adversarial attack, 
it is much more challenging. We categorize previous works into black-box and white-box attacking, depending on whether the attacker is aware of the gradients of the victim model. 
For black-box attacking, previous works usually replace words or characters in the original inputs 
guided by some restrictions to maintain the original semantics, including heuristic priors~\citep{jin2019bert,li2018textbugger,ren2019generating}, semantic similarity checking~\citep{li2020bert,cer2018universal}, extra language models~\citep{alzantot2018generating} or a combination of them~\citep{jin2019bert,li2020bert,zang2020word,morris2020reevaluating}.
For white box attacking, most previous works follow Fast Gradient Sign Method~(FGSM)~\citep{goodfellow2014explaining} to replace the original input with tokens that are able to drastically 
increase the prediction loss of
the victim model, either
at the char-level~\citep{ebrahimi2017hotflip}, word-level~\citep{wallace2019universal,papernot2016crafting},
or phrase-level~\citep{liang2017deep}.
In this paper, we focus on white-box attacking and
propose a unified framework that integrates the three atomic operations in a novel manner, by introducing and manipulating a blank token \texttt{[BLK]}.

\subsection{Non-Autoregressive NMT}
\label{sec:related-nat}

Non-autoregressive neural machine translation~(NAT)~\citep{gu2017non} is proposed to speedup the inference latency of autoregressive machine translation models, by generating target tokens in parallel while decoding. Among the follow-up works, mask-predict decoding~\citep{ghazvininejad2019mask} is shown effective by achieving comparable performance with autoregressive models while halving the inference latency~\citep{guo2020jointly}.
However, in the mask-predict model,
the target length requires to be predicted up front, making the results unstable and sensitive. 
The training-inference discrepancy 
is also a key problem~\citep{ghazvininejad2020semi}.

We show that our method can be viewed as an effective data augmentation method, which is able to improve 
the translation performance of the NAT model, as well as alleviating the problems mentioned above.

\section{Methodology}
\label{sec:metho}

\begin{figure*}[tb]
\centering
\includegraphics[width=0.8\linewidth]{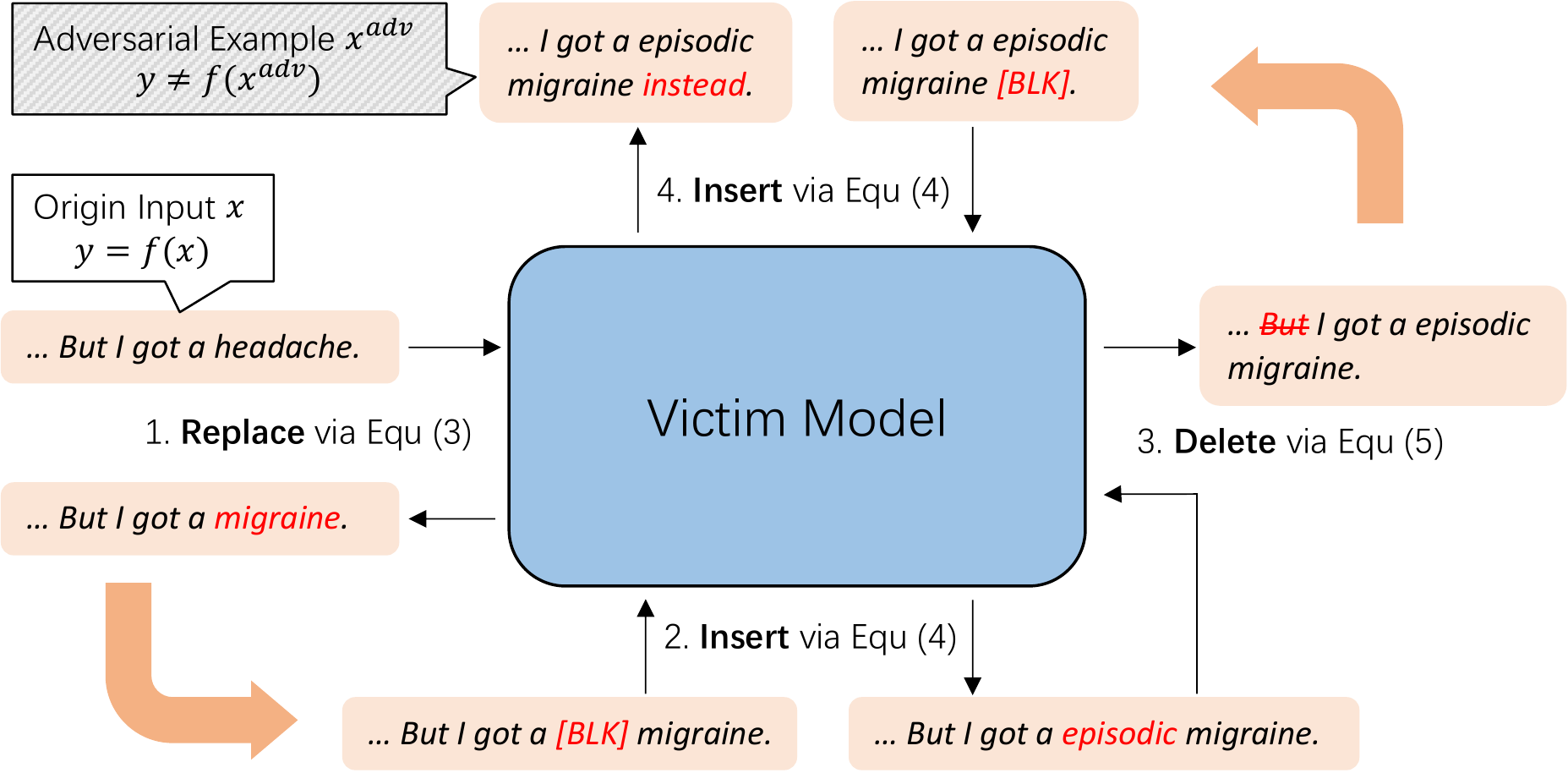}
\caption{An illustration of the attacking pipeline with the proposed insertion, deletion and replacement operations. We take the second half of the sample in Figure~\ref{fig:illus} as the example. $(x,y)$ is a pair of test sample where $x$ is the input while $y$ is the target, and $x^{\textrm{adv}}$ is the generated adversarial sample through $4$ attacking steps.}
\label{fig:pipe}
\vspace{-10pt}
\end{figure*}

We introduce the proposed Variable-Length textual adversarial Attack~(VL-Attack) method in this section. We start with the problem definition.

\paragraph{Problem Definition}
Given a pre-trained model such as BERT, after fine-tuning it on the training set of a downstream task $\mathcal{T}_{\textrm{train}}$, we get the victim model $f(\cdot)$ which will be evaluated on the test set $\mathcal{T}_{\textrm{test}}$. 
We denote the task-specific loss function while fine-tuning as $L(y, f(x))$.
Given a test sample $(x,y) \in \mathcal{T}_{\textrm{test}}$ where $x$ is the input sequence and $y$ is its target, there should be $f(x)=y$. 
While attacking, we aim at generating an adversarial example $x_{\textrm{adv}}$ that satisfies:
\begin{equation}
\label{equ:def}
f(x^{\textrm{adv}}) \neq y \text{ and } \operatorname{sim}(x,x^{\textrm{adv}}) > \theta,
\end{equation}
where $\operatorname{sim}(\cdot)$ indicates the function that measures the semantic similarity between the original inputs and adversarial examples, where $\theta$ is the threshold. 

\subsection{Variable-Length Attack}
\label{sec:metho-attack}
Our adversarial attack 
consists of multiple steps, and in each step, the attack is conducted by an atomic operation selected randomly from \textit{replacement}, \textit{insertion} and \textit{deletion}.
Formally, denote $\mathcal{M}(x)=x^{\textrm{adv}}$ as the mapping function from the original input to the adversarial example, 
it can be defined as:
\begin{align}
\mathcal{M}(x) &= M_k \odot M_{k-1} \cdots \odot M_1(x), \\
M_{i} &\in \{ \textrm{insertion}, \textrm{deletion}, \textrm{replacement} \}, \nonumber
\end{align}
where $k$ is the number of operation steps as well as the Levenshtein distance between $x$ and $x^{\textrm{adv}}$.

Each step of our attack is achieved by introducing and leveraging a special blank token \texttt{[BLK]}. While fine-tuning, given a training sample $(x,y) \in \mathcal{T}_{\textrm{train}}$, we randomly insert a blank token into the given input $x=(x_1,x_2,...,x_n)$, resulting $x^\prime=(x_1,...,x_{i-1},\texttt{[BLK]}, x_i,...,x_n)$. Then we fine-tune the model on $(x^\prime,y)$ instead of the original sample $(x,y)$, i.e., train the model with $L(y, f(x^\prime))$. In this way, the \texttt{[BLK]} token is learned to serve as a blank space as its occurrence will not affect the semantic information of the original sample.
With the help of the blank token, we introduce the proposed three atomic operations as follows.
Note that we conduct adversarial attacks while inference, i.e., $(x,y)\in \mathcal{T}_{\textrm{test}}$.

\paragraph{Replacement}
The replacement operation is inspired by the FGSM method~\citep{goodfellow2014explaining} which leverages the first-order approximation of the loss function. For the $i$-th token $x_i$ in the original input $x$, we replace it with the token that most drastically increases the loss function in the vocabulary,
\begin{equation}
\label{equ:rep}
x^{\textrm{adv}}_i = \argmax_{x_j\in V_{x_i}} (e(x_j) - e(x_i))^\intercal \nabla_{x_i} L(y, f(x)),
\end{equation}
where $e(x_i)$ indicates the embedding of $x_i$, $V_{x_i}=\operatorname{top\_k} \{ P_{\textrm{lm}}(x_j|x_{<i}, x_{>i}) \}$ is a subset of the whole vocabulary $V$, which is refined by a language model $P_{\textrm{lm}}$. In our setting, we directly use the victim BERT model as the language model for simplicity.

\paragraph{Insertion}
The insertion operation can be divided into two steps. Firstly, we construct a sample $x^\prime$ that is semantically identical to $x$ but a token longer, i.e., $|x^\prime|=|x|+1$ and $\operatorname{sim}(x,x^{\textrm{adv}}) > \theta$.
We achieve this by inserting a \texttt{[BLK]} token into $x$ as in fine-tuning.
Then, the token to insert is determined in a similar way as in the replacement operation, 
\begin{align}
\label{equ:insert}
&x^{\textrm{adv}}_{\texttt{BLK}} = \\
&\argmax_{x_j\in V_{x_{\texttt{BLK}}}} (e(x_j) - e(x_{\texttt{BLK}}))^\intercal \nabla_{x_{\texttt{BLK}}} L(y, f(x^\prime)), \nonumber
\end{align}
where $x_{\texttt{BLK}}$ represents the inserted \texttt{[BLK]} token, and $V_{x_{\texttt{BLK}}}$ is the vocabulary refined by the language model similar as above. 

In replacement, the position of the attacked token is determined by computing Equation~(\ref{equ:rep}) over the whole input sequence $x$ and selecting the position that increases the loss the most. Similarly, we iteratively insert the \texttt{[BLK]} token into each position and compute Equation~(\ref{equ:insert}) to determine where to conduct the insertion operation.
Both operations can be efficiently computed in parallel.

\paragraph{Deletion}
We propose the deletion operation based on the following intuition: the token that is more dissimilar to the \texttt{[BLK]} token will be more important to the prediction result.
Specifically, we 
quantify the importance of tokens in $x$ with the help of \texttt{[BLK]} token,
\begin{equation}
\label{equ:del}
\alpha_i = (e(x_i) - e(x_{\texttt{BLK}}))^\intercal \nabla_{x_i} L(y, f(x)),
\end{equation}
where $\alpha_i$ indicates the importance of the $i$-th token $x_i$. Then the deleted token $x_{\textrm{del}}$ is selected among all tokens, where $\textrm{del}= \argmax_{j\in |x|} \alpha_j$ and $|x|$ denotes the sequence length of the input $x$. In this way, the deleted token is selected jointly by its distance to the \texttt{[BLK]} token as well as the increase of loss in the gradient space.

Based on the proposed operations, we introduce the pipelines of our method when attacking NLU tasks as follows. 
We provide an illustration in Figure~\ref{fig:pipe}.
We aim at attacking the model with minimum number of modifications. Given an input $x$ and its label $y$, we iteratively apply the attack steps introduced above until obtaining a successful attack (i.e., $f(x^{\textrm{adv}}) \neq y$) or reaching the upper bound of attack operations $N=\lfloor \lambda \cdot |x| \rfloor$, where $\lambda$ is a hyper-parameter controlling the trade-off between the overall attack success rate and the modification rate.
In each step, we measure the semantic similarity between $x$ and $x^{\textrm{adv}}$ by Universal Sentence Encoder~\citep{cer2018universal} which encodes two sentences into a pair of fixed-length vectors and calculate the cosine similarity between them. This is a common practice in previous works~\citep{jin2019bert,li2020bert}. We skip this attack step if the similarity is less than the threshold $\theta$.

\subsection{Adversarial Training on NAT Models}
\label{sec:metho-nat}
Our method can be naturally applied for adversarial training to enhance the robustness of the victim model.
Specifically, we evaluate our method on NAT with mask-predict decoding~\citep{ghazvininejad2019mask}. 
Given a bilingual training pair $(x,y)$, the model is trained as a conditional masked language model,
\begin{equation}
\label{equ:cmlm_loss}
L_{\textrm{NAT}}(y^{m}| y^{r},x)= 
- \sum_{t=1}^{|y^m|} \log P(y^{m}_{t}|y^{r},x),
\end{equation}
where $y^m$ are masked tokens and $y^r$ are residual target tokens, following the same masking strategy proposed in BERT~\citep{devlin2018bert}.
While decoding, different from traditional autoregressive models which dynamically determine the target length with the \texttt{[EOS]} symbol, the NAT model needs to predict the target length at first, 
i.e., modeling $P(\left |y \right| | x)$, 
making the translation result sensitive to the predicted length.
In addition, as pointed out by~\citet{ghazvininejad2020semi}, there exists the discrepancy between training and inference
in mask-predict decoding, since the observed decoder inputs are correct golden targets in training but are not always correct in inference.

By providing variable-length adversarial examples, our method can alleviate these problems by adversarially training the NAT model.
The attack procedure is different from that on NLU tasks in two aspects. Firstly, we obtain adversarial examples by conducting a predefined number of attack steps as 
there does not exist a precise definition of a ``successful attack'' in machine translation.
Secondly, we conduct attacks jointly on the encoder input $x$ as well as the decoder input $y^r$, resulting the adversarial training loss $L_{\textrm{NAT}}(y^{m}| (y^r)^{\textrm{adv}},x^{\textrm{adv}})$.
By doing so, the model observes plausible decoder inputs other than the golden targets, thus alleviating the training-inference discrepancy. 
In addition, by keeping the targets and feeding the model with variable-length encoder and decoder inputs, the model robustness regrading the length prediction will also be enhanced.

\section{Experiments}
We conduct experiments
in two parts to evaluate the proposed method. On NLU tasks, we verify the adversarial attack performance of our method. On the task of non-autoregressive machine translation, we explore the potential of our method for adversarial training.
We start with NLU tasks.

\subsection{Adversarial Attack on NLU}

\begin{table*}[tb]
\centering
\small
\caption{Results of adversarial attacks against various fine-tuned BERT models.
For BERT-Attack and TextFooler, we directly copy the scores reported by~\citep{li2020bert}, and we re-implement the word-level HotFlip when attacking BERT based on the their code.
For the attacked accuracy~(AttAcc) and perturbed ratio~(Perturb\%), the lower the better, while for the semantic similarity~(Sim), the higher the better.
}
\vspace{-10pt}
\setlength{\tabcolsep}{6pt}
\begin{tabular}{@{} l ccc m{3pt} ccc @{}}
\toprule
{} & \multicolumn{3}{c}{\textbf{Yelp} (OriAcc = 96.8)} && \multicolumn{3}{c}{\textbf{IMDB} (OriAcc = 90.4)}
\\\midrule
\textbf{Model}  &
\textbf{AttAcc}$\downarrow$ & \textbf{Perturb\%}$\downarrow$ &  \textbf{Sim}$\uparrow$ & &
\textbf{AttAcc}$\downarrow$ & \textbf{Perturb\%}$\downarrow$ &  \textbf{Sim}$\uparrow$
\\\midrule
 BERT-Attack  &
 $5.1$ &  $4.1$ & $0.77$ &&
 $11.4$ & $4.4$ &  $0.86$
 \\
 TextFooler  &
 $6.6$ &  $12.8$ &  $0.74$ &&
 $13.6$ &  $6.1$ &  $0.86$
 \\
 HotFlip  &
 $9.2$ &  $12.3$ &  $0.63$ &&
 $8.2$ &  $2.7$ &  $0.84$
 \\
 \midrule
 VL-Attack &
 $\textbf{4.8}$ &  $\textbf{3.7}$ &  $\textbf{0.83}$ &&
 $\textbf{3.5}$ &  $\textbf{1.3}$ &  $\textbf{0.87}$
\\\midrule\midrule

{} & \multicolumn{3}{c}{\textbf{SNLI} (OriAcc = 90.0)} && \multicolumn{3}{c}{\textbf{MNLI} (OriAcc = 85.1)}
\\\midrule
\textbf{Model}  &
\textbf{AttAcc}$\downarrow$ & \textbf{Perturb\%}$\downarrow$ &  \textbf{Sim}$\uparrow$ & &
\textbf{AttAcc}$\downarrow$ & \textbf{Perturb\%}$\downarrow$ &  \textbf{Sim}$\uparrow$
\\\midrule
 BERT-Attack  &
 $11.8$ &  $10.9$ & $0.48$ &&
 $9.9$ & $8.4$ &  $0.62$
 \\
 TextFooler  &
 $12.4$ &  $26.0$ &  $0.50$ &&
 $17.5$ &  $20.9$ &  $0.61$
 \\
 HotFlip  &
 $6.3$ &  $11.0$ &  $0.54$ &&
 $8.0$ &  $9.4$ &  $0.61$
 \\
 \midrule
 VL-Attack &
 $\textbf{4.8}$ &  $\textbf{9.5}$ &  $\textbf{0.56}$ &&
 $\textbf{5.3}$ &  $\textbf{7.3}$ &  $\textbf{0.66}$
\\\bottomrule
\end{tabular}
\label{tab:attack-results}
\vspace{-10pt}
\end{table*}

\subsubsection{Experimental Setup}
In NLU tasks, we fine-tune the pre-trained BERT base model on downstream classification datasets, and take it as the victim model. 
We mainly follow the settings in previous works~\citep{jin2019bert,li2020bert} to construct a fair comparison.
\paragraph{Datasets}
We evaluate our method on benchmark NLU tasks including two sentiment classification datasets, i.e., YELP and IMDB, as well as two natural language inference datasets, i.e., MNLI and SNLI.
\begin{itemize} [leftmargin=*]
    \setlength{\itemsep}{0pt}
    \setlength{\parskip}{0pt}
\item \textbf{YELP}: A document-level sentiment classification dataset on restaurant reviews. We process them into a polarity classification task following~\citep{zhang2015character}.
\item \textbf{IMDB}: A binary document-level sentiment classification dataset on movie reviews\footnote{\url{https://datasets.imdbws.com/}}.
\item \textbf{MNLI}: A multi-genre natural language inference dataset 
    consisting of premise-hypothesis pairs, and the task is to determine the relation of the hypothesis to the premise, i.e., entailment, neutral, or contradiction~\citep{williams2017broad}. We evaluate on the matched set.
\item \textbf{SNLI}: A natural language inference dataset from the Stanford language inference task~\citep{bowman2015large}.
\end{itemize}
For all tasks, we evaluate our method on the $1$k test samples provided by~\citep{alzantot2018generating} and also utilized in~\citep{jin2019bert,li2020bert}, which are randomly selected from the corresponding test sets of each task.
We tokenize and segment each word into wordpiece tokens w.r.t the vocabulary of the pre-trained BERT model, and we conduct attacks 
at the wordpiece level.
We set the upper bound of perturbations $\lambda=30\%$ 
and the semantic similarity threshold $\theta=0.5$ for all tasks.

\paragraph{Baselines}
We compare our method with two state-of-the-art black-box adversarial attack methods including TextFooler~\citep{jin2019bert} and BERT-Attack~\citep{li2020bert}. 
We strictly follow the settings in~\citep{li2020bert} and directly copy the results of TextFooler and Bert-Attack as reported in their papers.
We also consider a white-box baseline HotFlip~\citep{ebrahimi2017hotflip}, which 
conducts attacks mainly based on the replacement operation. We consider the word-level variant of their model, and obtain their results when attacking BERT by re-implementing the model based on their code\footnote{https://github.com/AnyiRao/WordAdver}.

\paragraph{Evaluation}
We validate the performance of our model with both automatic metrics and human evaluation. The automatic evaluation includes the following metrics.
The \textbf{Original Accuracy~(OriAcc)} indicates the original model prediction accuracy without adversarial attacks, and the \textbf{Attacked Accuracy~(AttAcc)} indicates the model performance on the adversarial examples generated by the attack methods. Lower AttAcc represents the more successful attacks. The \textbf{Perturbed Ratio~(Perturb\%)} is the average percentage of perturbed tokens when conducting attacks. Intuitively, the attack is more efficient and semantic preserving with less perturbations and higher attacked accuracy. Finally, we report the \textbf{Semantic Similarity~(Sim)} between the original inputs and adversarial examples measured by the averaged Universal Sentence Encoder~(USE)~\citep{cer2018universal} score.

As for human evaluation, we follow the settings of~\citep{jin2019bert} to measure the grammaticality, semantic similarity as well as label consistency of the original sentences and the adversarial examples. We randomly sample $100$ sentences from the test set of MNLI as well as IMDB and generate adversarial examples by our method, and then invite three human experts to annotate the results, who are all native speakers with university-level education backgrounds. The grammaticality is scored from $1$ to $5$, and the semantic similarity is determined by measuring whether the generated example is similar/ambiguous/dissimilar to the original sentence, scoring as $1$/$0.5$/$0$ respectively. In addition, the label consistency between the pair of an original and generated sentence is also collected based on human classification.
We report the results of our method and the HotFlip baseline. Human annotators are blind to method identities.

\subsubsection{Automatic Evaluation}

The results of adversarial attack on NLU tasks are listed in Table~\ref{tab:attack-results}. The proposed VL-Attack outperforms the compared state-of-the-art black-box baselines as well as the replacement based white-box baseline in all settings.
On sentiment classification tasks which are easier to attack, our method drastically drops the accuracy of the victim model by 
more than $95\%$ with only editing less than $4\%$ tokens.
Our method achieves similar performance with editing less than $10\%$ tokens on the harder natural language inference tasks.

Comparing with HotFlip, the baseline based on the replacement operation, our method is able to achieve more successful attacks with less perturbations as well as maintaining higher semantic similarity, which illustrates that more natural adversarial examples can be 
obtained by introducing the insertion and deletion operations.

\subsubsection{Ablation Study}

\begin{table}[tb]
\centering
\small
\caption{The ablation study of the proposed VL-Attack with naively implemented insertions and deletions. Results are obtained on the IMDB dataset, and we keep 
the same settings as in Table~\ref{tab:attack-results}.}
\vspace{-10pt}
\begin{tabular}{l|ccc}
  \toprule
  \textbf{Model} & \textbf{AttAcc} & \textbf{Perturb\%} & \textbf{Sim} \\
  \midrule
  VL-Attack & $3.5$ & $1.3$ & $0.87$ \\
   ~~+ Naive Insert & $5.6$ & $3.9$ & $0.77$ \\
   ~~+ Naive Delete & $7.9$ & $5.3$ & $0.80$ \\
  \bottomrule
  \end{tabular}
  \label{tab:ablation}
\vspace{-5pt}
\end{table}

\begin{table}[tb]\setlength{\tabcolsep}{3pt}
\centering
\small
\caption{Human evaluation results, where ``/'' indicates the results are only applicable for adversarial samples.}
\vspace{-10pt}
\begin{tabular}{cc|ccc}
  \toprule
  \multicolumn{2}{c|}{\textbf{Dataset}} & \textbf{Grammar} & \textbf{Semantic} & \textbf{Consistency} \\
  \midrule
  \multirow{3}*{\textbf{MNLI}} &Original  &$4.71$ & / & / \\
         & HotFlip &$3.98$ & $0.75$ & $0.60$ \\
         & VL-Attack &$4.30$ & $0.86$ & $0.77$ \\
  \midrule
   \multirow{2}*{\textbf{IMDB}} & Original & $4.88$ & / & / \\
         & HotFlip & $4.02$ & $0.78$ & $0.73$ \\
         & VL-Attack & $4.57$ & $0.89$ & $0.85$ \\
  \bottomrule
  \end{tabular}
  \label{tab:human}
\vspace{-10pt}
\end{table}

\begin{table*}[ht]\setlength{\tabcolsep}{4pt}\small
\caption{Case studies on the MNLI dataset of the proposed VL-Attack method.  ``P'' and ``H'' indicate premise and hypothesis respectively. ``Origin'' indicates the original input sample.
    Red words indicate replacements and underlined words indicate insertions.
    }
    \vspace{-10pt}
    \begin{tabular}{l|l|l}
        \toprule 
        \textbf{Method}&\textbf{Content}&\textbf{Label} \\
        \midrule

        \multirow{3}*{Origin} &
        P: Sandstone and granite were the materials used to build the Baroque church of Bom Jesus, &\multirow{3}*{Contradiction}\\ 
        & famous for its casket of St. Francis Xavier's relics in the mausoleum to the right of the altar. & \\
        & H: St. Francis Xavier's relics were never recovered, unfortunately. & \\

        \cmidrule{2-3} 
        \multirow{3}*{VL-Attack} &
        P: Sandstone and granite were the materials used to build the Baroque church of Bom Jesus, &\multirow{3}*{Neutral}\\ 
        & famous for its casket of St. Francis Xavier's relics in the mausoleum to the right of the altar. & \\ 
        & H: St. Francis Xavier's relics were never \textcolor{red}{\textit{recapture}}, \textcolor{red}{\textit{however}}. &\\
        
        \midrule
        
        \multirow{3}*{Origin} &
        P: Julius Caesar's nephew Octavian took the name Augustus;  & \multirow{3}*{Contradiction} \\
        & ~~~~~Rome ceased to be a republic, and became an empire. &\\ 
        & H: Rome never ceased to be a republic, and did not become an empire. & \\

        \cmidrule{2-3} 
        \multirow{3}*{VL-Attack} &
        P: Julius Caesar's nephew Octavian took the name Augustus; &\multirow{3}*{Entailment} \\
        & ~~~~~Rome \textcolor{red}{\textit{\ul{not}}} \textcolor{red}{\textit{come}} to be a republic, and became an empire. &\\
        & H: Rome never ceased to be a republic, and did not become an empire. &\\
        
    \bottomrule
    \end{tabular}
    \label{tab:samples}
\vspace{-10pt}
\end{table*}

We achieve the insertion and deletion operations by introducing and leveraging a special blank token.
The benefits of 
doing so can be concluded as twofold.
Firstly, the \texttt{[BLK]} token is trained as a blank space to the model, which can be treated as a placeholder that does not change the original semantics of the input. In this way, we ensure that the inserted token is selected 
based on the original context of the input, which cannot be achieved if other tokens are used as the placeholder
(e.g., using existing special tokens such as \texttt{[MASK]} or duplicating the current token)
while inserting.

Secondly, determining the importance of tokens to efficiently conduct  attacks correspondingly, is an important problem in textual adversarial attack~\citep{li2020bert}. Previous works achieve this by greedily running predictions~\citep{li2020bert} or leveraging external tools~\citep{liang2017deep}, which lacks efficiency.
Here, the importance of each token can be easily determined by its distance to \texttt{[BLK]} following Equation~(\ref{equ:del}) as a byproduct of introducing the blank token.

Here, we verify the efficacy of the blank token by naively implementing the insertion and deletion operations. Specifically, we change the \texttt{[BLK]} token to 
the \texttt{[MASK]} token
and then conduct insertion following the same method described in Section~\ref{sec:metho-attack}.
For deletion, we randomly delete a token of the input.

We test these naive baselines on the IMDB dataset, and results are shown in Table~\ref{tab:ablation}.
We can find that naive insertion results in lower semantic similarity, indicating that
the \texttt{[MASK]} token will
change the context of the original input while insertion, and thus maintaining less original semantic information.
Naive deletion achieves worse attacked accuracy with perturbing more tokens, which indicates the effectiveness and efficiency of the proposed blank token when determining the importance of different tokens and conducting attacks.

\subsubsection{Human Evaluation}
The human evaluation results are listed in Table~\ref{tab:human}, where we report the average scores of all annotators.
Consistent with the results of the automatic evaluation, the adversarial examples generated by our method are both grammatically and semantically closer to the original samples than the replacement based baseline. 
In addition, a majority of generated examples preserve the same labels as the original samples according to human annotators, showing that our attacking method is indistinguishable to human judgement most of the time.

\subsubsection{Case Study}
In Table~\ref{tab:samples}, we provide several adversarial examples generated by our method on the MNLI dataset.
Perturbations are marked as red.
The generated examples are generally semantic consistent with the origin input, as well as indistinguishable in human knowledge.

\subsubsection{Adversarial Training on NLU Tasks}

\begin{table}[tb]
\centering
\small
\caption{Results of fine-tuning a pre-trained BERT model with adversarial training on NLU tasks.}
\begin{tabular}{l|cccc}
  \toprule
  \textbf{Model} & \textbf{YELP} & \textbf{IMDB} & \textbf{SNLI} & \textbf{MNLI} \\
  \midrule
  Origin & $96.8$ & $90.4$ & $90.0$ & $85.1$ \\
   ~~+ VL-Attack & $96.4$ & $91.2$ & $90.1$ & $84.7$ \\
  \bottomrule
  \end{tabular}
  \label{tab:adv_nlu}
\end{table}

We also conduct additional investigations of adversarial training on NLU tasks to verify whether the generated adversarial examples preserve the original labels. 
Specifically, we first generate adversarial examples on the training set of each task, which are then concatenated with the original training set to fine-tune a pre-trained BERT model. And we evaluate the model on clean test sets,
where the model performance can
be drastically dropped if most adversarial examples are inconsistent with their original labels.

Results are listed in Table~\ref{tab:adv_nlu}.
With adversarial training, the model performs comparably with the original performance (change of accuracy varies from $-0.4$ to $+0.8$) on clean test sets, 
showing that most generated adversarial examples are consistent with the original label and will not hurt the model performance.

\subsection{Adversarial Training on NAT}
In this section,
we utilize our method for adversarial training on NAT tasks to alleviate some key problems in NAT models as discussed in Section~\ref{sec:metho-nat}. 
The adversarial training results on NLU tasks are also provided in the supplementary material.

\subsubsection{Experimental Setup}
\paragraph{Dataset and Model Configurations}
We conduct experiments on benchmark machine translation datasets including IWSLT14 German$\rightarrow$English~(IWSLT14 De-En)\footnote{https://wit3.fbk.eu/}, WMT14 English$\leftarrow$German translation~(WMT14 En-De)\footnote{https://www.statmt.org/wmt14/translation-task}, and WMT16 English$\leftrightarrow$Romanian~(WMT16 En-Ro)\footnote{https://www.statmt.org/wmt16/translation-task}.
For IWSLT14, we adopt the official split of train/valid/test sets. For WMT14 tasks, we utilize \texttt{newstest2013} and \texttt{newstest2014} as the validation and test set respectively. For WMT16 tasks, we use \texttt{newsdev2016} and \texttt{newstest2016} as the validation and test set.
We tokenize the sentences by Moses~\citep{koehn2007moses} and segment each word into subwords using Byte-Pair Encoding~(BPE)~\citep{sennrich2015neural}, resulting in a $32$k vocabulary shared by source and target languages.
On WMT tasks, the model architecture is akin to the base configuration of the Transformer model~\citep{vaswani2017attention}~($d_{\textrm{hidden}}=512$, $d_{\textrm{FFN}}=2048$, $n_{\textrm{layer}}=6$, $n_{\textrm{head}}=8$). We use a smaller configuration on the IWSLT14 De-En task~($d_{\textrm{hidden}}=256$, $d_{\textrm{FFN}}=512$, $n_{\textrm{layer}}=5$, $n_{\textrm{head}}=4$).

We choose the mask-predict~\citep{ghazvininejad2019mask} model as the victim NAT model.
We use the Transformer base configuration on WMT tasks and a smaller configuration on the IWSLT14 De-En task.
Following mask-predict, we utilize sequence-level knowledge distillation~\citep{kim2016sequence} on the training set of the WMT14 En-De task to provide less noisy and more deterministic training data for NAT models~\citep{gu2017non}. And we use raw datasets for other tasks.
We train the model on 1/8 Nvidia P100 GPUs for IWSLT14/WMT tasks.
For evaluation, we report the tokenized BLEU scores~\citep{papineni2002bleu} measured by \texttt{multi-bleu.perl}.

\paragraph{Adversarial Training Pipeline}
We first pre-train a mask-predict model, on which we execute adversarial attacks on the training set following the process described in Section~\ref{sec:metho-attack}, except that we conduct a fixed number (randomly sampled from $1$ to $15\%$ of the sequence length) of attacking steps as a ``successful attack'' in NLG tasks is not well defined.
Then we fine-tune the model on the concatenation of the original training set as well as the attacked training set.
As for baselines, we compare our method with HotFlip, a white-box attacking method mainly based on the replacement operation. 
We consider its word-level variant.

\begin{table}[tb]\setlength{\tabcolsep}{4pt}
\centering
\small
\caption{The BLEU scores of mask-predict with adversarial training and 
baselines on the IWSLT14 De-En, WMT14 En-De and WMT16 En-Ro tasks. The results of baselines are obtained by our implementation.
}
\vspace{-10pt}
\begin{tabular}{l|ccc}
\toprule
 \multicolumn{1}{c|}{} & \multicolumn{1}{c}{\textbf{IWSLT14}} & \multicolumn{1}{c}{\textbf{WMT14}} & \multicolumn{1}{c}{\textbf{WMT16}} \\
\textbf{Models} & \multicolumn{1}{c}{De$-$En} & \multicolumn{1}{c}{En$-$De} & \multicolumn{1}{c}{En$-$Ro} \\
\midrule
Transformer  & $32.59$ & $28.04$ & $34.13$  \\
\midrule
Mask-Predict     & $31.71$ & $27.03$     & $33.08$      \\
\quad + HotFlip & $32.05$ & $26.91$     & $33.12$     \\
\quad + VL-Attack     & $\textbf{33.18}$ & $\textbf{27.55}$     & $\textbf{33.57}$   \\
\bottomrule
\end{tabular}
\label{tab:nat_bleu_results}
\vspace{-5pt}
\end{table}

\begin{table}[tb]
\centering 
\small
\caption{The BLEU scores of mask-predict when facing adversarial attacks. The results are obtained on the IWSLT14 German-English translation task.
}
\vspace{-10pt}
\begin{tabular}{l|cc}
\toprule
 \multicolumn{1}{c|}{} & \multicolumn{2}{c}{\textbf{IWSLT14 De-En}} \\
\textbf{Models} & \multicolumn{1}{c}{\textbf{dev}} & \multicolumn{1}{c}{\textbf{test}} \\
\midrule
Mask-Predict     & $32.38$ & $31.71$  \\
\quad Attacked by HotFlip     & $22.95$ & $21.01$ \\
\quad Attacked by VL-Attack     & $21.29$ & $19.47$ \\
\midrule
Mask-Predict + VL-Attack    & $33.84$ & $33.18$  \\
\quad Attacked by HotFlip     & $30.79$ & $29.85$ \\
\quad Attacked by VL-Attack     & $28.42$ & $27.13$ \\
\bottomrule
\end{tabular}
\label{tab:nat_adv_defense}
\vspace{-10pt}
\end{table}

\subsubsection{Results}
The main results are listed in Table~\ref{tab:nat_bleu_results}. Our method outperforms the original mask-predict model as well as the HotFlip baseline on all datasets. Specifically, we can find that HotFlip, the replacement based adversarial attack method, does not provide clear improvements over the original mask-predict model, indicating that fixed-length adversarial examples are not helpful for the training of NAT models.
On the contrary, the proposed VL-Attack is able to boost the performance of the mask-predict model by providing variable-length adversarial examples.

\subsubsection{Adversarial Defense}
Here, we verify that in addition to enhancing the model performance on regular test sets, our method can also help the model defend against adversarial attacks.
We attack the original mask-predict model as well as the model fine-tuned by our method on the validation and test set of the IWSLT14 German-English translation task, and we set the attacking steps as $3$.

Results are shown in Table~\ref{tab:nat_adv_defense}, from which we can
find that the proposed VL-Attack provides more critical attacks than HotFlip as more BLEU scores are dropped.
In addition, fine-tuning with adversarial examples generated by our method successfully promotes the robustness of the mask-predict model. Specifically, on the test set,
the drop of BLEU score has been decreased from $33.74\%$ to $10.04\%$ for HotFlip and from $38.60\%$ to $18.23\%$ for the proposed method, showing that the adversarial training method can help the model
defend both fixed-length and variable-length adversarial attacks.

\section{Conclusion}

In this paper, we propose a variable-length adversarial attack method on textual data, which is consisted of three atomic operations including replacement, insertion and deletion. We integrate them into a unified framework by introducing and leveraging a special blank token, which is trained to serve as a blank space to the victim model, and therefore can be utilized as a placeholder while inserting and measuring the importance of tokens while deleting. 
We verify the effectiveness of the proposed method by attacking a pre-trained BERT model on natural language understanding tasks, and then show that our method is beneficial to non-autoregressive machine~(NAT) translation, a task that is sensitive to the input lengths, when treated as a data augmentation method. Our method is able to boosts the translation performance as well as the robustness of adversarial attacks of the NAT model.
In the future, we will try to
extend our method to the black-box setting, by alternating gradient-based replacement to heuristic approaches, or leveraging imitation learning~\citep{wallace2020imitation}.

\bibliography{emnlp2021}
\bibliographystyle{acl_natbib}

\end{document}